\documentclass[10pt,twocolumn,letterpaper]{article}

\usepackage{iccv}
\usepackage{times}
\usepackage{epsfig}
\usepackage{graphicx}
\usepackage{amsmath}
\usepackage{amssymb}
\usepackage[toc,page]{appendix}
\usepackage{booktabs}
\usepackage[ruled,lined,linesnumbered]{algorithm2e}
\usepackage{algorithmic}
\usepackage{makecell}
\usepackage{multirow}

\usepackage[breaklinks=true,bookmarks=false]{hyperref}

\iccvfinalcopy 

\def\iccvPaperID{****} 

\begin{document}

\title{UGAN: Untraceable GAN for Multi-Domain Face Translation}
\author{
Defa Zhu\textsuperscript{\rm 1}, Si Liu \textsuperscript{\rm 2}, Wentao Jiang \textsuperscript{\rm 2}, Chen Gao\textsuperscript{\rm 1}, Tianyi Wu\textsuperscript{\rm 3}, Qiangchang Wang\textsuperscript{\rm 4}, Guodong Guo\textsuperscript{\rm 3}\\ 
\small{
\textsuperscript{\rm 1}Chinese Academy of Sciences, 
\textsuperscript{\rm 2}Beihang University, 
\textsuperscript{\rm 3}Baidu Research, 
\textsuperscript{\rm 4}West Virginia University} \\
{\tt\small \{zhudefa, gaochen\}@iie.ac.cn}, {\tt\small\{liusi, jiangwentao\}@buaa.edu.cn}\\ {\tt\small \{wutianyi01, guoguodong01\}@baidu.com}, {\tt\small qw0007@mix.wvu.edu} 
}
\maketitle

\begin{abstract}
  The multi-domain image-to-image translation is a challenging task where the goal is to translate an image into multiple different domains. The target-only characteristics are desired for translated images, while the source-only characteristics should be erased. However, recent methods often suffer from retaining the characteristics of the source domain, which are incompatible with the target domain. To address this issue, we propose a method called Untraceable GAN, which has a novel source classifier to differentiate which domain an image is translated from, and determines whether the translated image still retains the characteristics of the source domain. Furthermore, we take the prototype of the target domain as the guidance for the translator to effectively synthesize the target-only characteristics. The translator is learned to synthesize the target-only characteristics and make the source domain untraceable for the discriminator, so that the source-only characteristics are erased. Finally, extensive experiments on three face editing tasks, including face aging, makeup, and expression editing, show that the proposed UGAN can produce superior results over the state-of-the-art models. The source code will be released. 

\end{abstract}
   

\section{Introduction}
  
\begin{figure}[t]
 \begin{center}
  \includegraphics[width=0.48\textwidth]{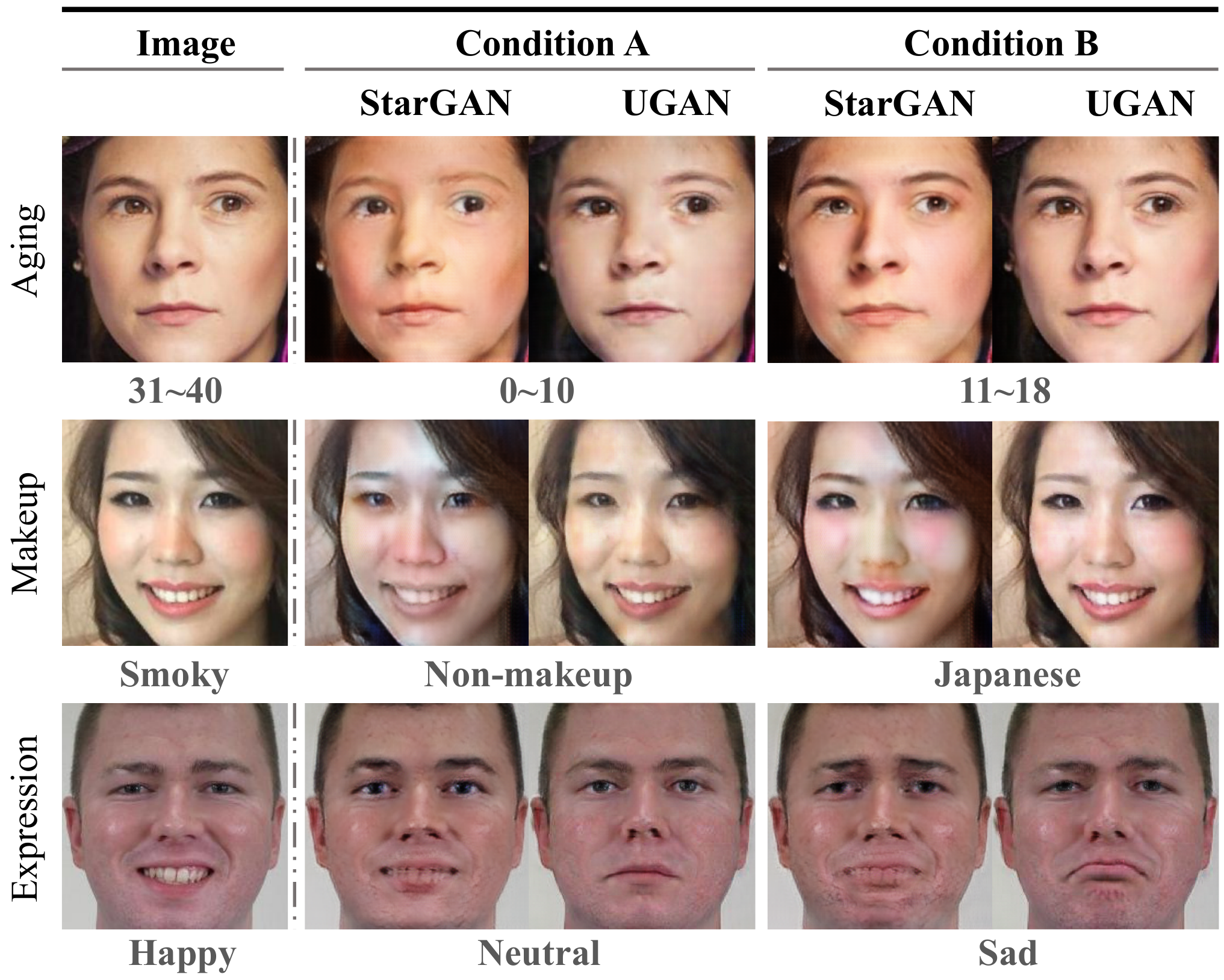}
 \end{center}
 \caption{The phenomenon of source retaining in image translation. 
 The first column shows input images. 
 In the face aging case, when changing a face from $31\sim40$ to $0\sim10$, the result of StarGAN still looks like an adult while that of UGAN is more like a child with big eyes and smooth skin. When translating the face to $11\sim18$, the result of UGAN also looks more like a juvenile. Similar observations can be made in expression and makeup editing tasks.
 }
 \label{fig:first_fig}
\end{figure}

Multi-domain image-to-image translation \cite{choi2017stargan} refers to image translation among multiple domains, where each domain is characterized by different attributes. 
For example, the face aging task, with age groups as domains, aims to translate a given face into other age groups using a single translator.  
As shown in Figure \ref{fig:first_fig} Row 1, the input face image is translated into different age groups. 

Although prior works \cite{choi2017stargan,zhao2018modular,he2019attgan} have made significant progress, the translated results still suffer from retaining the characteristics of the source domain (incompatible with the target domain), which is the so-called \emph{phenomenon of source retaining}. 
As illustrated in Figure \ref{fig:first_fig} Row 1, 
when StarGAN translates a female face from the age group 31$\sim$40 to 0$\sim$10, the translated image still looks like an adult. 
In makeup editing shown in Figure~\ref{fig:first_fig} Row 2, StarGAN fails to eliminate the eye shadows in makeup removing. 
For expression editing, as shown in Figure \ref{fig:first_fig} Row 3, the results of StarGAN show visible teeth shadows around the mouth region. 

\begin{figure}[t]
 \begin{center}
  \includegraphics[width=0.45\textwidth]{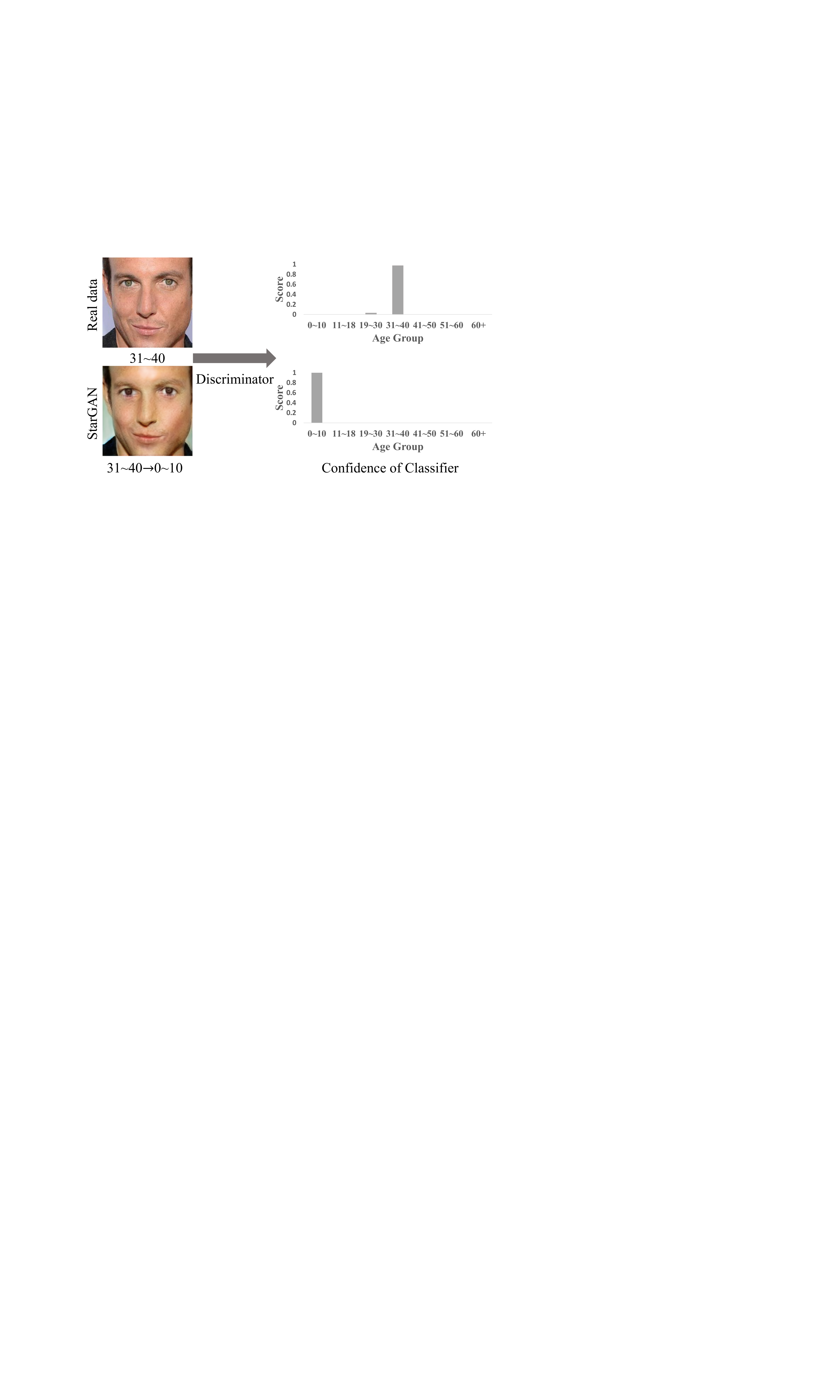}
 \end{center}
 \caption{The domain classifier of the discriminator in StarGAN is easily deceived on face aging task. First row: Given an adult face within 31$\sim$40 age group from the test set, the domain classifier of the discriminator can successfully recognize the corresponding age. Second row: The adult face is translated into 0$\sim$10 years old, and the translated face heavily retains adult characteristics including beard and expression wrinkles. However, the classifier still doesn't identify the incompatible characteristics and is completely fooled by the translated face.}
 \label{fig:story}
\end{figure}

The reason for \emph{phenomenon of source retaining} is that the explicit and effective mechanisms to erase the characteristics of the source domain have not been explored in the prior works. 
Most of them just simply apply a domain classifier, which is only trained to recognize the domain class of real data, to guide the image translation. However, the domain classifier is not sensitive to the non-qualified synthesized image containing incompatible characteristics. 
As shown in Figure \ref{fig:story} Row 1, the discriminator correctly judges an adult face to be within 31$\sim$40 age group. Translating the adult face into a child face (0$\sim$10), the translated face heavily retains adult characteristics, e.g., beard and expression wrinkles (Figure \ref{fig:story} Row 2). However, the discriminator judges it to be within 0$\sim$10 age group with the confidence of $1$. 
That is, the synthesized image containing incompatible characteristics has almost no punishment from the domain classifier, which results in the \emph{phenomenon of source retaining}. 

  
To tackle the problem of source retaining, we propose a new method untraceable GAN (UGAN), which introduces \textit{untraceable constraint} and \textit{prototype injection}. 
The untraceable constraint is employed to encourage the translator to erase all the source-only characteristics and synthesize certain target-only ones. 
As shown in Figure \ref{fig:story}, the process of an image from 31$\sim$40 years old (source domain) translated to 0$\sim$10 years old (target domain), 
the beard and wrinkles (source-only characteristics) need to be erased, while a smooth skin and round face (target-only characteristics) should be synthesized.
To endow the proposed UGAN with the above capabilities, a discriminator is trained to \textit{track} which domain the synthesized image is \textit{translated from}, while the translator is trained to make the source domain of the synthesized image being untraceable for the discriminator.
Furthermore, To effectively synthesize the target-only characteristics, we take the prototype \cite{Kemelmacher2014Illumination} of the target domain as the guidance for the translator. The prototype is a statistic of the target domain, which aims to provide the essential characteristics, like the round face of 0$\sim$10 years old domain. 


Our contributions include: 
\begin{itemize}
 \item To the best of our knowledge, this is the first work to present the phenomenon of source retaining in multi-domain image-to-image translation, and propose a novel UGAN to explicitly erase the characteristics of the source domain for improving the image translation.
 \item A novel source classifier is introduced to differentiate which domain an image is translated from, and determines whether the translated image still retains the characteristics of the source domain.  
 \item The propose UGAN is the first work to take the target prototype into the translator for synthesizing the target domain characteristics. 
 \item Extensive qualitative and quantitative experiments are conducted for three face editing tasks that demonstrate the superiority of our proposed UGAN.  
\end{itemize}



\section{Related Work}

In this section, we give a brief review on three aspects related to our work: Generative Adversarial Network, Conditional GANs and Image-to-Image Translation.

{\bf Generative Adversarial Networks} (GANs) \cite{Goodfellow2014Generative} are popular generative models that employ adversarial learning between a generator and discriminator to synthesize the realistic data, which have gained astonishing successes in many computer vision tasks, such as image-to-image translation \cite{isola2017image}, domain adaptation \cite{liu2018cross} and super-resolution \cite{ledig2017photo}. 
In this work, the proposed UGAN enjoys the adversarial learning \cite{arjovsky2017wasserstein,gulrajani2017improved}, which approximately minimizes the Wasserstein distance between the synthesized distribution and real distribution. 

{\bf Conditional GANs} \cite{mirza2014conditional} are variants of GANs, which aim to controllably synthesize examples under the given condition. 
Many prior works focus on generating samples under different forms of conditions, such as category label in the form of one-hot code \cite{mirza2014conditional} or learnable parameters \cite{miyato2018spectral}, and text with word embedding \cite{zhang2017stackgan}, etc. 
Different from these works, for synthesizing the required characteristics, we introduce the prototype of the condition to provide prior information, where the prototype is one of the statistics of the target domain.  
{\bf Image-to-Image Translation}
is first defined in pix2pix~\cite{isola2017image}, which is improved from various aspects, such as 
skip connection for maintaining useful original information \cite{zhang2018generative,liu2017face,pumarola2018ganimation}, cascade training from coarse to fine \cite{wang2017high,dekel2018sparse}, extra relevant data \cite{choi2017stargan}, buffer of history fake image~\cite{shrivastava2017learning}, multi-discriminator~\cite{wang2017high}, 3D technology~\cite{yao20183d}, variational sampling~\cite{zhu2017toward,esser2018variational}. 
If the translator only models directed translation between two domains, $C\cdot (C-1)$ translators are required among $C$ domains. A single conditional translator for multi-domain translation is seriously demanded. Thus we focus on multi-domain translation with such a single translator. 
The current multi-domain image translation methods \cite{choi2017stargan,zhao2018modular,he2019attgan} using the vanilla one-hot condition for the translator, without considering the information contained in each domain.
We are the first to adopt the statistics of each domain as a condition of the translator to efficiently inject the essential characteristics. 
Furthermore, the prior methods apply the domain classifier for condition constraints. However, limited by the classifier, they often suffer from the \emph{phenomenon of source retaining}. 
Thus, we change the role of this auxiliary classifier in UGAN and make it classify which source domain the given datum is translated from, instead of classifying which domain the given datum is sampled from. 
\begin{figure*}[t]
  \begin{center}
    \includegraphics[width=1\textwidth]{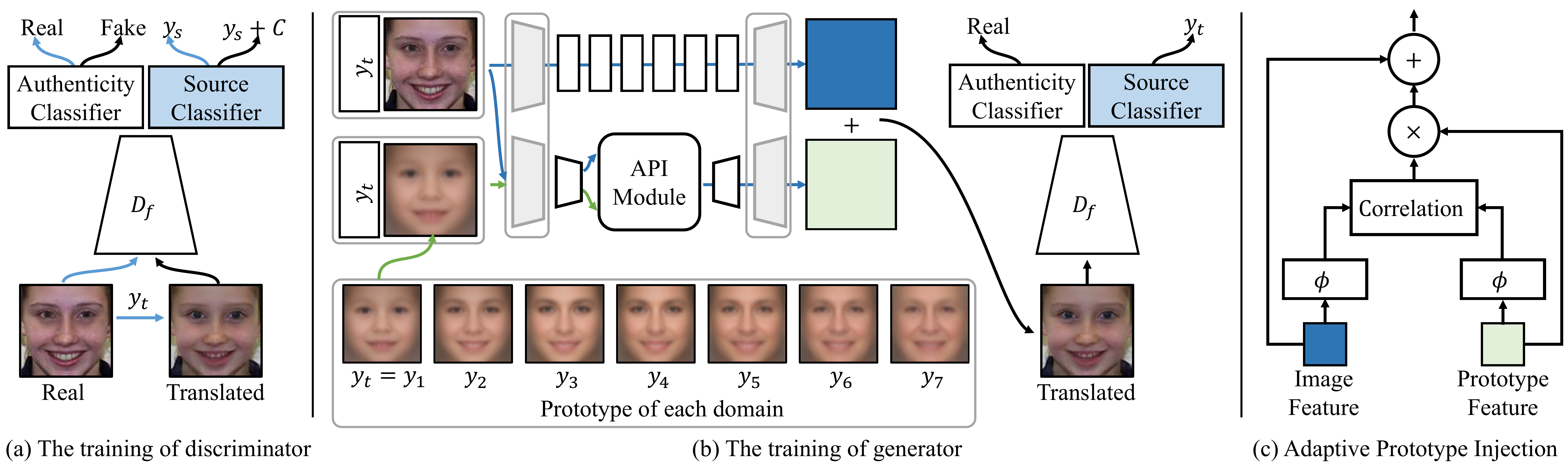}
  \end{center}
  \caption{An overview of UGAN. a) The discriminator D should not only distinguish whether the input sample is real or fake but also determine which domain the sample is translated from. b) Translator G is trained to fool D by synthesizing realistic images of the target domain. We take the average image as a ``prototype'' to inject prior information of the target domain. c) The adaptive prototype injection is introduced for injecting essential characteristics of the prototype into the source image.}
  \label{fig:framework}
\end{figure*}

\section{Our Approach}
\label{sec:method}

The framework of UGAN is shown in Figure \ref{fig:framework}. The input image and the target condition are fed into the translator $G$. 
The discriminator $D$ has two heads: one head is named as the authenticity classifier to distinguish whether the input sample is real or fake; the other is called the source classifier, aiming to determine which domain the sample is translated from, where the real data are supposed to be translated from themselves. 
For erasing source-only characteristics and synthesizing the target-only characteristics, translator $G$ is trained to fool the source classifier of $D$ to believe that the synthesized image is translated from the target domain. 
Moreover, to effectively synthesize the target characteristics, we introduce the ``prototype'' of the target domain and inject it into the translated image. 

For convenience, we then introduce the used mathematical annotations. 
Discriminator $D$ here contains two heads including the authenticity classifier $D_a$ and the source classifier $D_s$, where $D_a$ and $D_s$ share the same feature extraction module $D_f$. $D_a(D_f(\cdot))$ and $D_s(D_f(\cdot))$ are abbreviated as $D_a(\cdot)$ and $D_s(\cdot)$ respectively. $\{x_s, y_s\}$ is a sample pair from the source domain, where $x_s$ represents the image and $y_s$ is its label. By feeding the image $x_s$ and the target label $y_t$ into $G$, it produces $\tilde{x}_{s\rightarrow{t}}=G(x_s,y_t)$. 
We use $q(x,y)$ to denote the joint distribution of image $x$ and domain label $y$. $q(x)$ and $q(y)$ are the marginal distribution of images and labels, respectively. 

\subsection{Untraceable Constraint}


To tackle the problem of source retaining, the source classifier $D_s$ is trained to classify which domain image $x$ is {translated} from. 
For an real image-label pair $\{x_s,y_s\}$, we regard $x_s$ as translated from domain $y_s$ to domain $y_s$. Since $D_s$ aims to classify where an image is translated from, the real datum $x_s$ should be classified into $y_s$, meaning $x_s$ is {translated from} domain $y_s$. The synthesized image $\tilde{x}_{s\rightarrow{t}}$ should be classified into $y_s$, meaning $\tilde{x}_{s\rightarrow{t}}$ is translated from $y_s$. 
Translator $G$ is trained to fool $D_s$ to classify $\tilde{x}_{s\rightarrow{t}}$ into the $y_t$. 
In this way, $G$ is trained to make the source domain of $\tilde{x}_{s\rightarrow{t}}$ is untraceable and the target domain characteristics are injected to $\tilde{x}_{s\rightarrow{t}}$. 
The adversarial training is formulated as follows:
\begin{equation}
\begin{array}{*{20}{l}}
L_{D}^{S_1}=-E_{(x_s,y_s)}[logD_s(x_s,y_s)] \\
-\lambda_{u} E_{(x_s,y_s),y_t}[logD_s(G(x_s,y_t),y_s)],
\end{array}
\label{equ:dec-D}
\end{equation}
\begin{equation}
L_{G}^{S_1}=-\lambda_{u} E_{(x_s,y_s),y_t}[logD_s(G(x_s,y_t),y_t)],
\label{equ:dec-G}
\end{equation}
where $\lambda_{u}$ is the penalty coefficient of source retaining. For space limit, $E_{(x,y){\sim}q(x,y)}[\cdot]$ and $E_{y{\sim}q(y)}[\cdot]$ are abbreviated as $E_{(x,y)}[\cdot]$ and $E_{y}[\cdot]$, respectively. 

Note that the $\tilde{x}_{s\rightarrow{t}}$ should be injected with certain target-only characteristics. 
Recall that in Eq.~(\ref{equ:dec-G}), $G$ is trained to fool $D_s$ to classify $\tilde{x}_{s\rightarrow{t}}$ into $y_t$. 
However, the class $y_t$ here is not pure that mixed with the characteristics of $x_t$ and synthesized data $\tilde{x}_{t\rightarrow{?}}$. 
Refer to Eq.~(\ref{equ:dec-D}), the source classifier $D_s$ treats the real sample $x_t$ sampled from $y_t$ and fake sample $\tilde{x}_{t\rightarrow{?}}$ translated from $y_t$ as the same class $y_t$. 
To accurately synthesize the characteristics of the target domain, the number of categories of $D_s$ is augmented as $2C$. The first $C$ categories are real data and those sampled (translated) from the corresponding domain. The latter $C$ categories are fake data, and those translated from the corresponding domain. $y_s+C$ means input datum is fake and translated from $y_s$. In addition, the translator $G$ is trained to fool $D_s$ to classify $\tilde{x}_{s\rightarrow{t}}$ into the $y_t$ category. The untraceable constraint conducted via optimizing the following:
\begin{equation}
\begin{array}{*{20}{l}}
L_{D}^{S_2}=-E_{(x_s,y_s)}[logD_s(x_s,y_s)] \\
-\lambda_{u} E_{(x_s,y_s),y_t}[logD_s(G(x_s,y_t),y_s+C)],
\end{array}
\label{equ:dec-D-2}
\end{equation}
\begin{equation}
L_{G}^{S_2}=-\lambda_{u} E_{(x_s,y_s),y_t}[logD_s(G(x_s,y_t),y_t)].
\label{equ:dec-G-2}
\end{equation} 
In this process, $D_s$ is trained to identify whether $\tilde{x}_{s\rightarrow{t}}$ is a fake image and the source domain. $G$ is trained to approximate the true untraceable translator. 

\subsection{Prototype Injection}
The statistics of the target domain can provide guidance information for image translation. Refer to Figure \ref{fig:framework} (b), the average image of each age group shows the essential characteristics, like round face and flat nose characteristics of age group 1 (0$\sim$10). 
Thus, We leverage the statistics of the target domain to further inject the essential characteristics of the target domain into the translated image, where we call the statistic, containing the essential characteristics, as the ``prototypes'' following the classic aging method \cite{Kemelmacher2014Illumination}. 
However, the posture of the source image and target prototype may be misaligned. Thus, concatenating or summing up the image feature and prototype feature will hurt the performance. 
To naturally inject these essential characteristics, we design an adaptive prototype injection (API) module inspired by non-local operation \cite{wang2018non,vaswani2017attention}. 

Refer to Figure \ref{fig:framework} (c), the injection process of API is formulated as follows: 
\begin{equation}
  \begin{array}{*{20}{l}}
  \mathcal{A}_{ij} = \frac{
    exp( 
      \phi (f^x_i)
      \cdot 
      \phi (f^p_j)
      )
    }{
      \sum_{\forall k}
      exp(
        \phi (f^x_i)
        \cdot 
        \phi (f^p_k)
        )
    }, \\
  f^{inject}_{i} = f^x_i + \sum_{\forall j}\mathcal{A}_{ij} \cdot f^p_j, 
  \end{array}
  \label{equ:WGANgp-D}
\end{equation}
where $f^x$ and $f^p$ are the feature maps of the source image and target prototype, respectively. $i$ is the index of an feature map position. $\phi$ is a linear mapping to reduce the dimension. 
Since the computation of the correlation matrix $\mathcal{A}$ is computationally expensive, we apply the API module on the low-resolution feature maps. 
To simultaneously maintain resolution and inject the prototype, the translator $G$ is designed with two parallel networks, with parameter sharing at both ends (gray color, Figure \ref{fig:framework}). For maintaining the resolution, one network is a common architecture in image translation \cite{Zhu2017Unpaired}. The other one applies the API module on the low-resolution feature maps. Finally, the outputs of these two networks are fused by element-wise sum to generate the translated image. 

\subsection{Objective Function}
\textbf{Authenticity constraint:} The adversarial loss of WGAN-gp \cite{gulrajani2017improved} is adopted to constrain the synthetic joint distribution to approximate the real distribution. 
\begin{equation}
\begin{array}{*{20}{l}}
L_{D}^{A}=&-E_{x_s}[D_a{(x_s)}] 
+E_{x_s,y_t}[D_a(G(x_s,y_t))] \\ 
&+{\lambda_{gp}}E_{\hat{x}}[(\left\|{\nabla}_{\hat{x}}[D_a(\hat{x})\right \|_{2}-1)^2],
\end{array}
\label{equ:WGANgp-D}
\end{equation}
\begin{equation}
L_{G}^{A}=-E_{x_s,y_t}[D_a(G(x_s,y_t))], 
\label{equ:WGANgp-G}
\end{equation}
where $\hat{x}=\alpha\cdot{x_s}+(1-\alpha)\cdot{G(x_s,y_t)}$, and $\alpha\sim{U(0,1)}$. The third term in Eq.~(\ref{equ:WGANgp-D}) is a gradient penalty term that enforces the discriminator as a 1-Lipschitz function.

\textbf{Cycle Consistency:} The input and output are regularized to satisfy the correspondence \cite{Zhu2017Unpaired}:
\begin{equation}
L_{G}^{C}=\lambda_{c} E_{(x_s,y_s),y_t}[\|G(G(x_s,y_t),y_s)-x_s\|_1].S
\label{equ:cycle-G}
\end{equation}

\textbf{Overall loss function:} $D$ and $G$ are trained by optimizing
\begin{equation}
L_{D}^{U}=L_{D}^{A}+L_{D}^{S},
\label{equ:full-D-SE}
\end{equation}
\begin{equation}
L_{G}^{U}=L_{G}^{C}+L_{G}^{A}+L_{G}^{S},
\label{equ:full-G-SE}
\end{equation}
where $L^{S}$ could be $L^{S_1}$ or $L^{S_2}$. The Eq. (\ref{equ:full-D-SE}) and Eq. (\ref{equ:full-G-SE}) are optimized alternatively. 


\begin{table*}[t]
  \centering
  \caption{Intra FID on CFEE dataset.}
  \setlength{\tabcolsep}{1.3pt}
  \begin{tabular}{llllllllllllllllllllllll}
    \toprule
    Method & A & B & C & D & E & F & G & H & I & J & K & L & M & N & O & P & Q & R & S & T & U & V & Mean \\
    \midrule
    StarGAN & 52.1 & 52.6 & 61.4 & 51.5 & 55.9 & 64.1 & 57.8 & 54.1 & 42.6 & 52.5 & 61.7 & 69.3 & 55.2 & 51.9 & 55.0 & 63.2 & 68.0 & 60.6 & 69.9 & 61.0 & 59.1 & 61.3 & 58.2 \\
    UGAN$^\dag$ & 44.5 & 44.4 & 53.8 & 46.9 & 49.9 & 59.0 & 47.8 & 48.5 & 37.7 & 43.2 & 52.9 & 59.1 & 53.4 & 50.4 & 53.0 & \textbf{45.6} & 56.6 & 52.7 & 48.4 & 49.0 & 47.0 & 46.4 & 49.6 \\
    UGAN$^\ddag$ & 42.8 & 45.3 & 48.3 & 43.7 & 47.5 & \textbf{56.0} & 43.6 & 44.7 & 37.6 & 41.4 & 47.4 & \textbf{52.4} & 42.9 & 43.1 & 48.5 & 46.3 & \textbf{52.1} & \textbf{46.6} & 46.6 & \textbf{46.8} & 45.4 & \textbf{45.0} & 46.1 \\
    UGAN & \textbf{39.7} & \textbf{40.8} & \textbf{47.9} & \textbf{39.7} & \textbf{43.8} & 57.6 & \textbf{42.0} & \textbf{43.1} & \textbf{33.5} & \textbf{40.8} & \textbf{45.7} & 55.4 & \textbf{40.8} & \textbf{40.9} & \textbf{46.9} & \textbf{43.4} & 52.3 & 47.3 & \textbf{44.9} & 48.1 & \textbf{42.0} & 48.5 & \textbf{44.8} \\
    \bottomrule
  \end{tabular}
  \label{tab:intra FID on CFEE}
\end{table*}

\begin{table}[t]
  \centering
  \caption{Intra FID on face aging dataset.}
  \setlength{\tabcolsep}{0.4pt}
  \begin{tabular}{lcccccccr}
    \toprule
    method & 0$\sim$10 & 11$\sim$18 & 19$\sim$30 & 31$\sim$40 & 41$\sim$50 & 51$\sim$60 & 60+ & Mean \\
    \midrule
    CAAE & 63.8 & 64.1 & 67.6 & 69.8 & 75.9 & 78.7 & 87.2 & 72.4 \\
    C-GAN & 83.9 & 60.7 & 54.9 & 54.7 & 57.4 & 61.7 & 70.2 & 63.4 \\
    StarGAN & 59.9 & 38.2 & 29.9 & 41.4 & 37.3 & 40.0 & 46.9 & 41.9 \\
    \midrule \midrule
    UGAN$^\dag$ & \textbf{42.0} & 33.6 & 25.2 & 27.2 & 28.9 & 34.4 & 40.4 & 33.1 \\
     
    UGAN$^\ddag$ & 44.0 & 29.5 & 21.1 & 21.3 & 25.4 & 28.2 & 34.7 & 29.2 \\
    UGAN & 42.7 & \textbf{28.4} & \textbf{19.4} & \textbf{18.9} & \textbf{22.8} & \textbf{26.9} & \textbf{32.5} & \textbf{27.4} \\
    \bottomrule
  \end{tabular}
  \label{tab:intra FID on aging}
\end{table}

\begin{table}[t]
  \centering
  \caption{Intra FID on MAKEUP-A5 dataset.}
  \setlength{\tabcolsep}{1.2pt}
  \begin{tabular}{lcccccc}
    \toprule
    method & Retro & Korean & Japanese & Non-makeup & Smoky & Mean \\
    \midrule
    StarGAN   & 110.9 & 86.2 & 74.5 & 84.4 & 91.9 & 89.6 \\
    UGAN$^\dag$ & 109.4 & 70.9 & 61.8 & 72.8 & 74.8 & 78.0 \\
    UGAN$^\ddag$ & 101.7 & \textbf{65.9} & 58.1 & 64.5 & \textbf{66.3} & 71.3 \\
    UGAN & \textbf{89.6} & 73.3 & \textbf{57.1} & \textbf{62.1} & 68.8 & \textbf{70.2} \\
    \bottomrule
  \end{tabular} 
  \label{tab:intra FID on makeup}
\end{table}

\section{Experiments}
\subsection{Datasets}

\begin{table*}[t]
  \centering
  \small
  \caption{AMT results on CFEE dataset($\%$).}
  \setlength{\tabcolsep}{2.5pt}
  \begin{tabular}{lllllllllllllllllllllll}
    \toprule
    Method & A & B & C & D & E & F & G & H & I & J & K & L & M & N & O & P & Q & R & S & T & U & V \\
    \midrule
    StarGAN & 17.0     & 8.7      & 13.7     & 14.7     & 29.0     & 13.7     & 33.3     & 14.7     & 22.0     & 19.0     & 18.7     & 22.7     & 31.3     & 29.0     & 17.0     & 13.3     & 18.3     & 27.0     & 37.7     & 19.0     & 19.3     & 7.3 \\
    UGAN & \textbf{83.0} & \textbf{91.3} & \textbf{86.3} & \textbf{85.3} & \textbf{71.0} & \textbf{86.3} & \textbf{66.7} & \textbf{85.3} & \textbf{78.0} & \textbf{81.0} & \textbf{81.3} & \textbf{77.3} & \textbf{68.7} & \textbf{71.0} & \textbf{83.0} & \textbf{86.7} & \textbf{81.7} & \textbf{73.0} & \textbf{62.3} & \textbf{81.0} & \textbf{80.7} & \textbf{92.7} \\
    \bottomrule
  \end{tabular}
  \label{tab:AMT on CFEE}
\end{table*}

\begin{figure*}[t]
  \begin{center}
    \includegraphics[width=1\textwidth]{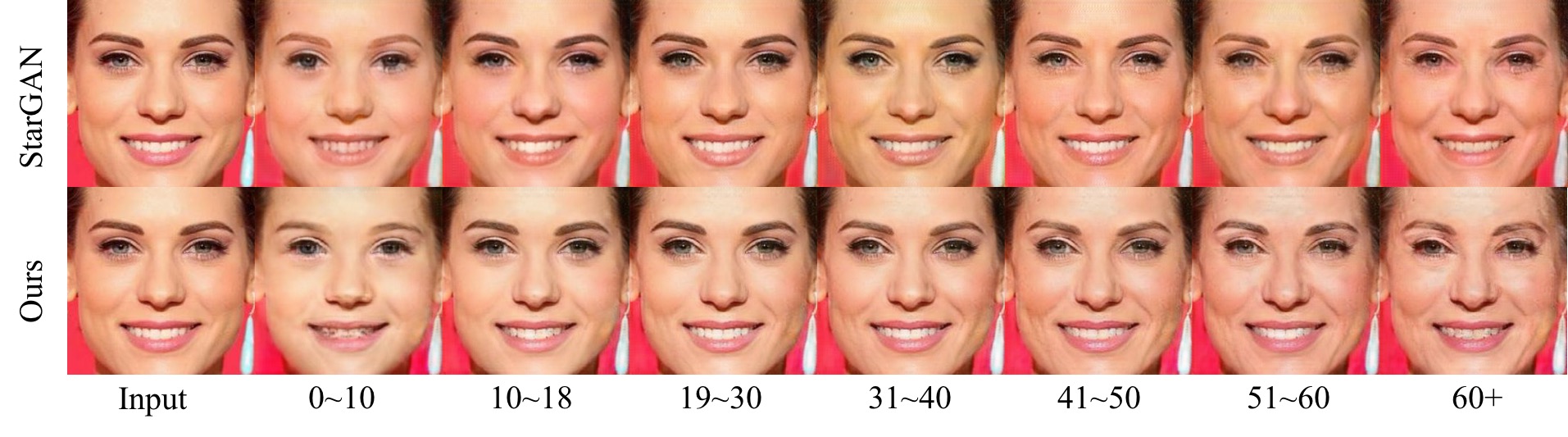}
  \end{center}
  \caption{Comparison of face aging synthesis results on the face aging dataset. }
  \label{fig:aging}
\end{figure*}

\begin{figure*}[t]
  \begin{center}
    \includegraphics[width=0.8\textwidth]{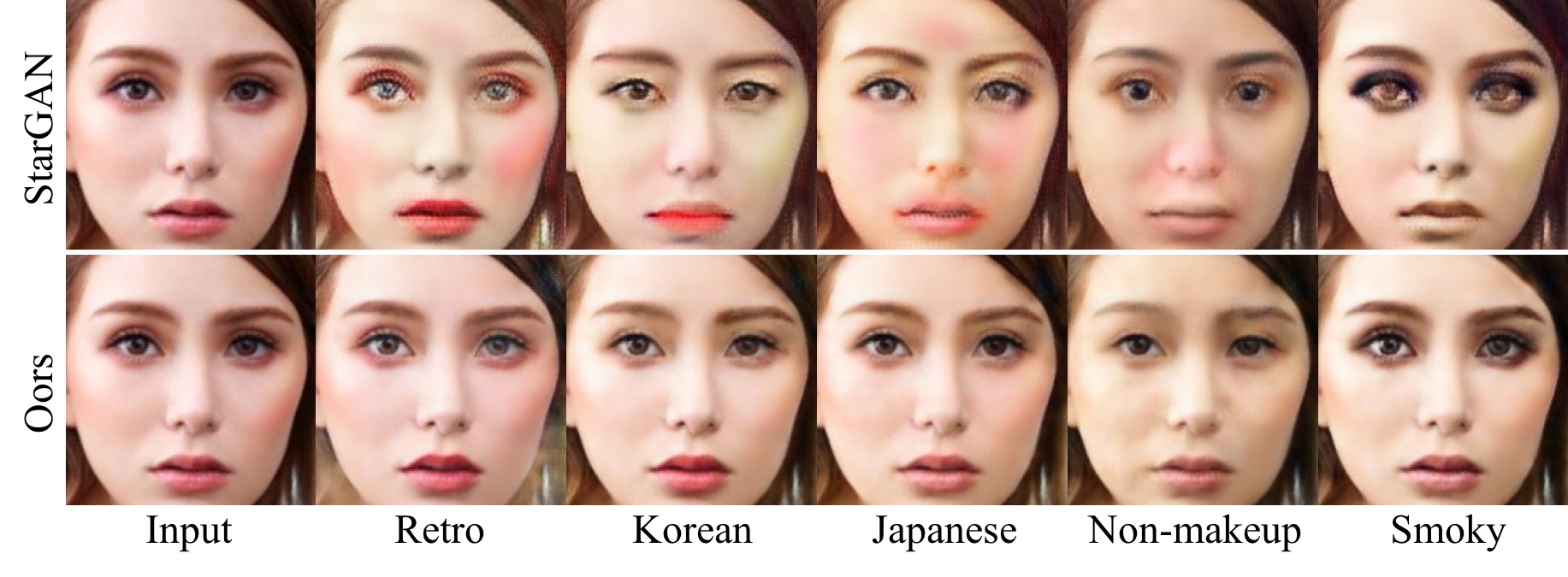}
  \end{center}
  \caption{Makeup synthesis results on the MAKEUP-A5. }
  \label{fig:makeup}
\end{figure*}

\begin{table}[t]
  \centering
  \setlength{\tabcolsep}{1pt}
  \caption{AMT results on face aging dataset ($\%$).}
  \begin{tabular}{llllllll}
    \toprule
    Method & 0$\sim$10 & 11$\sim$18 & 19$\sim$30 & 31$\sim$40 & 41$\sim$50 & 51$\sim$60 & 60+   \\
    \midrule
    StarGAN  & 13.0     & 34.3     & 34.3     & 42.3     & 39.7     & 10.7     & 13.0     \\
     
    UGAN & \textbf{87.0} & \textbf{65.7} & \textbf{65.7} & \textbf{57.7} & \textbf{60.3} & \textbf{89.3} & \textbf{87.0} \\
    \bottomrule
  \end{tabular} 
  \label{tab:AMT on aging}
\end{table}

\begin{table}[t]
  \centering
  \setlength{\tabcolsep}{2pt}
  \caption{AMT results on MAKEUP-A5 dataset ($\%$).}
  \begin{tabular}{lcccccc}
    \toprule
    Method & Retro & Korean & Japanese & Non-makeup & Smoky \\
    \midrule
    StarGAN & 28.7     & 40.3     & 21.3     & 16.7     & 30.7     \\
    UGAN & \textbf{71.3} & \textbf{59.7} & \textbf{78.7} & \textbf{83.3} & \textbf{69.3} \\
    \bottomrule
  \end{tabular} 
  \label{tab:AMT on makeup}
\end{table}


\begin{table}[t]
\centering
\setlength{\tabcolsep}{10.5pt}
\caption{Cosine similarity on hidden feature of ResNet-18 between source images and the corresponding translated images.}
\begin{tabular}{lcccc}
  \toprule
  \multirow{2}{*}{Method} & \multicolumn{4}{c}{Age Group Gap}                   \\ \cmidrule{2-5} 
              & $\geq3$ & $\geq4$ & $\geq5$ & $\geq6$ \\ \midrule
  StarGAN         & 0.757        & 0.742        & 0.745        & 0.719        \\ 
  UGAN         & \textbf{0.740}    & \textbf{0.714}    & \textbf{0.712}     & \textbf{0.696} \\
  \bottomrule    
  \end{tabular}
  \label{tab:similarity}
\end{table}

{\bf Face aging} dataset is collected by C-GAN \cite{liu2017face} including $15,030$ face images. Ages are divided into $7$ age groups including $0\sim{10}$, $11\sim{18}$, $19\sim{30}$, $31\sim{40}$, $41\sim{50}$, $51\sim{60}$ and $60+$. $10\%$ of the dataset is randomly selected as the test set, and the rest is the training set. 

{\bf MAKEUP-A5} is a makeup-labeled dataset \cite{li2018beautygan} containing $6,095$ aligned Asian woman faces with $5$ makeup categories including retro, Korean, Japanese, non-makeup and smoky. The training set contains $5,485$ images and the remaining is the test set. 

{\bf CFEE} is an expression dataset \cite{du2014compound} of $22$ expressions with $5,060$ images. The categories of facial expressions include (A) neutral, (B) happy, (C) sad, (D) fearful, (E) angry, (F) surprised, (G) disgusted, (H) happily surprised, (I) happily disgusted, (J) sadly fearful, (K) sadly angry, (L) sadly surprised, (M) sadly disgusted, (N) fearfully angry, (O) fearfully surprised, (P) fearfully disgusted, (Q) angrily surprised, (R) angrily disgusted, (S) disgustedly surprised, (T) appalled, (U) hatred and (V) awed. We randomly select $23$ identities ($506$ images) as the test set and use the other images for training. 
All images are aligned and resized to $256\times256$ resolution.

\subsection{Measurements}
{\bf Intra FIDs} \cite{heusel2017gans,miyato2018cgans,dowson1982frechet} on each domain and mean of them are used for evaluation. FID is a common quantitative measure for generative models, which measures the 2-Wasserstein distance between the two distributions $q$ and $p$ on the features extracted from InceptionV3 model. It is defined as \cite{dowson1982frechet}
\begin{equation}
\begin{array}{*{20}{l}}
F(q, p) = \left \|{\mu}_{q}-{\mu}_{p}\right \|_2^2 +\mathop{tr}(\sigma_{q}+\sigma_{p}-2(\sigma_{q}\sigma_{p})^{1/2}),
\end{array}
\end{equation}
where $q$ and $p$ are feature distributions of real data and synthesized data, $({\mu}_{q},\sigma_{q})$ and $({\mu}_{p},\sigma_{p})$ are the mean and the covariance of $q$ and $p$. The mean intra FID is calculated by
\begin{equation}
\begin{array}{*{20}{l}}
m{F}_{intra}(q, p) = \frac{1}{C} \sum_{i=1}^{C} F(q(\cdot|y_i), p(\cdot|y_i)),
\end{array}
\label{m_intra_fid}
\end{equation}
where $y$ is the domain label for the total $C$ domains. 

{\bf User studies by Amazon Mechanical Turk (AMT): } Given an input image, target domain images translated by different methods are displayed to the Turkers who are asked to choose the best one.

{\bf Cosine similarity:} For the face aging task, cosine similarity between the features of real images and the corresponding translated images is used to measure the degree of source retaining. Features are extracted by a ResNet-18 model \cite{he2016deep} trained on the same training set.  


\subsection{Implementation Details}
We perform experiments with three versions of our methods named as UGAN$^\dag$, UGAN$^\ddag$ and UGAN, where the methods with superscripts ($^\dag$ and $^\ddag$) mean adopting the same translator as StarGAN (without prototype), ``UGAN$^\dag$'' means adopting $L^{S_1}$ as untraceable constraint, while ``UGAN$^\ddag$'' adopting $L^{S_2}$. ``UGAN'' means the final method that adopting $L^{S_2}$ as an untraceable constraint and the proposed translator with an API module. 
For a fair comparison, our learning rate is fixed as $0.0001$, while the other hyper-parameters are kept the same as StarGAN. 
All experiments are optimized by Adam with $\beta_{1}=0.5$ and $\beta_{2}=0.999$. The discriminator is iterated $5$ times per iteration of the translator. All baselines and our methods are trained $200$ epochs. The mini-batch size is set to $16$. All images are horizontally flipped with a probability of $0.5$ as data augmentation.

{\bf Baselines:} StarGAN \cite{choi2017stargan} has shown the best performance than DIAT \cite{li2016deep}, CycleGAN \cite{Zhu2017Unpaired} and IcGAN \cite{perarnau2016invertible}. We, therefore, select StarGAN as our baseline to verify the superiority of our method. For the face aging task, we additionally compare two classic GAN-based methods of face aging, including CAAE \cite{zhang2017age} and C-GAN (without transition pattern network) \cite{liu2017face}.


\begin{figure*}[t]
  \begin{center}
    \includegraphics[width=1\textwidth]{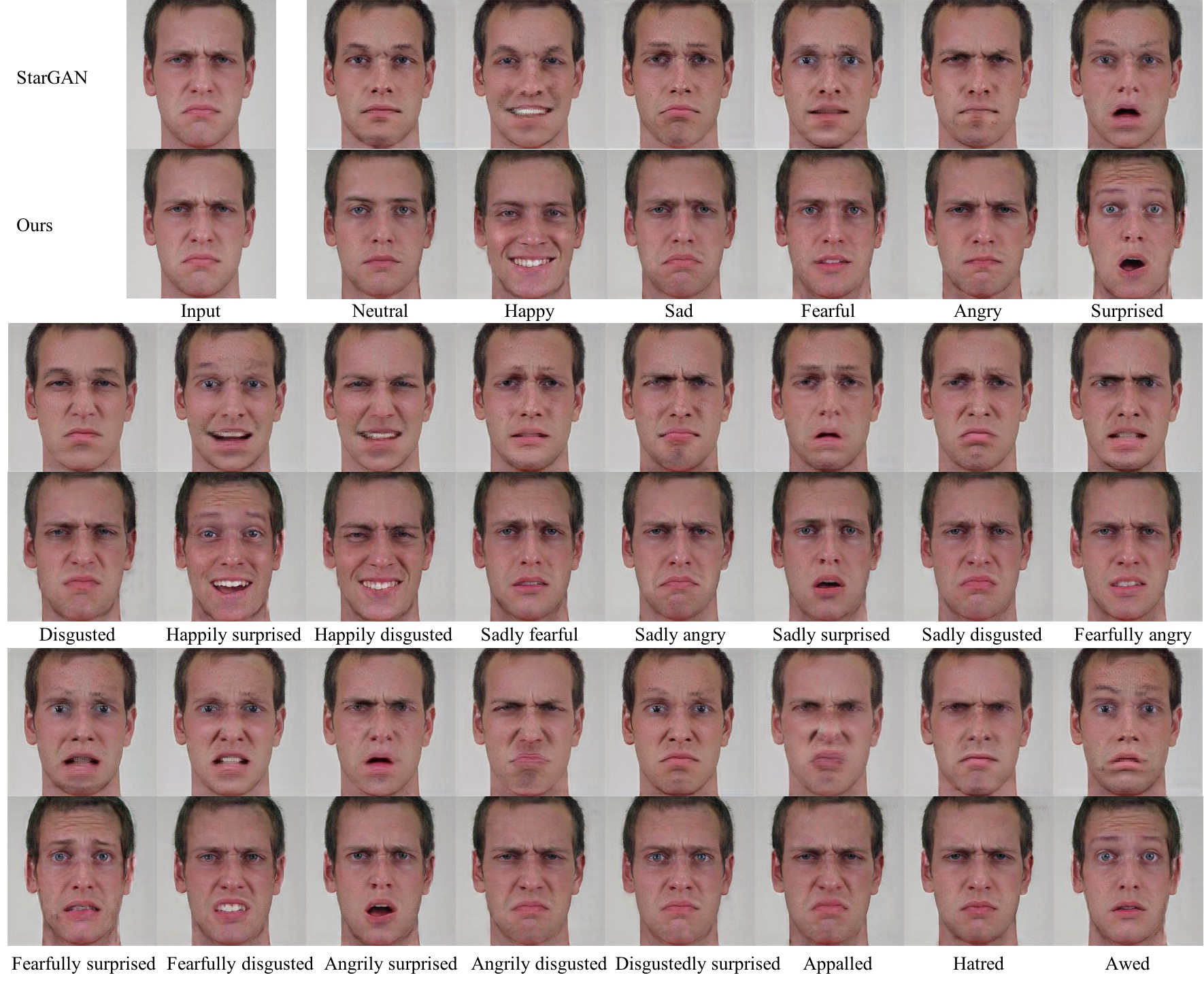}
  \end{center}
  \caption{Comparison of facial expression editing results on the CFEE dataset. }
  \label{fig:expression}
\end{figure*}

\subsection{Quantitative Experiments}
Given the domain label $y_i$, we traverse all images in the test set to generate fake images. All the synthetic images of each domain are adopted to calculate intra FID, while 
$300$ synthetic images of each domain are randomly sampled to be evaluated by AMT.

\textbf{Face aging:} The comparison of results on face aging dataset is shown in Table \ref{tab:intra FID on aging}. Face aging involves deformations and texture synthesis. For example, deformation, such as the face shape and eye size, are the main differences between babies and adults. Texture synthesis, like adding wrinkles, is also essential when translating a middle-aged man to a senior man.  
In Table \ref{tab:intra FID on aging}, both UGAN$^\dag$ and UGAN$^\ddag$ are significantly better than StarGAN on all age groups, where UGAN$^\ddag$ are better than UGAN$^\dag$.
The mean intra FID drops from $41.9$ (StarGAN) to $29.2$ (UGAN$^\ddag$). The relative drop is more than $30\%$. Furthermore, UGAN achieves the best performance with mean intra FID $27.4$.

\textbf{Makeup editing:} The comparison of results on MAKEUP-A5 dataset is shown in Table \ref{tab:intra FID on makeup}. Both texture and color need to be altered in makeup editing.
UGAN has the best performance in all categories. 
The mean intra FID has declined from $89.6$ (StarGAN) to $71.3$ (UGAN).

\textbf{Expression editing:} The comparisons on CFEE dataset are shown in Table \ref{tab:intra FID on CFEE}. The expression editing task aims to change the emotion of a face by deformation. The CFEE dataset contains $22$ kinds of fine-grained expressions, which makes the expression editing problem very challenging. From the results, 
we can conclude that UGAN again achieves the best performance. 
The mean intra FID is $58.2$ (StarGAN), $49.6$ (UGAN$^\dag$), $46.1$ (UGAN$^\ddag$), and $44.8$ (UGAN), respectively. It can be seen that the reduction is significant.

\textbf{AMT user studies:} 
For further evaluation, user studies are conducted on AMT \footnote{https://www.mturk.com/} to compare StarGAN and our method. Since UGAN outperforms UGAN$^\dag$ and UGAN$^\ddag$ for mean intra FID, only UGAN is compared. With datasets mentioned above, we synthesize $300$ pairs of images per domain by UGAN and StarGAN. All image pairs are shown to $102$ Turkers who are asked to choose the better one considering image realism and satisfaction of target characteristics. 
Table \ref{tab:AMT on CFEE}, \ref{tab:AMT on aging} and \ref{tab:AMT on makeup} show the percentage of our method beating StarGAN. 
For example, in Table \ref{tab:AMT on aging}, when changing a face to $0\sim10$ years old, StarGAN wins in $13.2\%$ cases while our method wins in $86.8\%$ cases. It again shows the advantages of our method when transforming a face into childhood. 
Generally, our method is better than StarGAN in every category of each dataset. 


\textbf{Tackling the phenomenon of source retaining:}
The effect of erasing source characteristics on face aging is shown in Table~\ref{tab:similarity}. 
A well-trained ResNet-18 (for age recognition) is adopted to extract features (the second last layer). We calculate average cosine similarity on the neural feature of all source images and translated image pairs from the test set. 
Intuitively, the smaller the similarity, the more thoroughly source characteristics are erased.
Since the images of adjacent age groups are similar, we only consider translation across a large age gap, e.g., across three age groups. In Table~\ref{tab:similarity}, we perform the experiments on multiple age group gaps, and the similarities of UGAN are smaller on all age group gaps.

\subsection{Qualitative Experiments}
The visualization results are shown in Figure \ref{fig:aging}, \ref{fig:makeup} and \ref{fig:expression}. More results are provided in supplementary material. 

\textbf{Face aging:} 
Results on the face aging dataset are shown in Figure \ref{fig:aging}. In the first example, an input image is a woman.
By comparing the results of $0\sim10$ years old (second column), our result has obvious childish characteristics, \emph{e.g.} round face, big eyes, and small nose, while the result of StarGAN does not look like a child.
Another example is the $60+$ years old case (last column). Our result has white hair, wrinkles, while StarGAN produces a middle-aged face.
These results show that UGAN can explicitly erase the characteristics of the source image by the source classifier in the discriminator.

\textbf{Makeup editing:} Two exemplary results on MAKEUP-A5 dataset are displayed in Figure \ref{fig:makeup}. For the first woman, by comparing the results of the second (retro) and last (smoky) columns, we find that blusher and eye shadows of UGAN are more natural, while StarGAN draws asymmetrical blusher and strange eye shadows. 
The result of UGAN is relatively natural when translating it to a non-makeup face. Therefore, we conclude that UGAN has learned the precise color and texture characteristics of different makeups.

\textbf{Expression editing:} Results on CFEE dataset are demonstrated in Figure \ref{fig:expression}. We have the following observations. First, UGAN can well edit $22$ kinds of fine-grained facial expressions. Also, UGAN captures the subtle differences between basic and compound expressions. For example, ``Happily surprised'' has bigger eyes and raising eyebrows compared to ``Happy''. Besides, the results of StarGAN under various expressions still retain the original expressions. For example, when changing the man from ``Hatred'' to ``Happy'', the result of StarGAN still has tight brows. 
Comparatively, UGAN can effectively synthesize the ``Happy'' expression by generating a grin and relaxed brows and erasing the tight brows.

\section{Conclusion}
The phenomenon of source retaining often occurs in the image-to-image translation task. 
To address it, the Untraceable GAN (UGAN) model has been proposed, where the discriminator estimates the source domain.
The translator $G$ is trained to fool the discriminator $D$ to believe that the generated data is translated from the target domain. 
In this way, the source domain of the synthesized image is untraceable.
In addition, we have further presented the prototype of each domain and inject it into the translated image to generate the target characteristics. 
Extensive experiments on three tasks have proven the significant advantages of our method over the state-of-the-art StarGAN. 

The source retaining phenomenon is common in various fields, where the UGAN idea may be widely used to alleviate the issue. 
For example, language translation \cite{lample2017unsupervised} often preserves the grammatical structure of the source language. UGAN may serve as a solution to improve translation quality. Furthermore, the prototype injection idea also can be introduced to the universal conditional generation. We plan to study these ideas in-depth and apply them to broader applications.

\small{
\bibliographystyle{ieee}
\bibliography{egbib}

\begin{thebibliography}{10}\itemsep=-1pt

\bibitem{arjovsky2017wasserstein}
M.~Arjovsky, S.~Chintala, and L.~Bottou.
\newblock Wasserstein gan.
\newblock {\em arXiv:1701.07875}, 2017.

\bibitem{choi2017stargan}
Y.~Choi, M.~Choi, M.~Kim, J.-W. Ha, S.~Kim, and J.~Choo.
\newblock Stargan: Unified generative adversarial networks for multi-domain
  image-to-image translation.
\newblock In {\em CVPR}, 2018.

\bibitem{dekel2018sparse}
T.~Dekel, C.~Gan, D.~Krishnan, C.~Liu, and W.~T. Freeman.
\newblock Sparse, smart contours to represent and edit images.
\newblock In {\em CVPR}, 2018.

\bibitem{dowson1982frechet}
D.~Dowson and B.~Landau.
\newblock The fr{\'e}chet distance between multivariate normal distributions.
\newblock {\em MA}, 1982.

\bibitem{du2014compound}
S.~Du, Y.~Tao, and A.~M. Martinez.
\newblock Compound facial expressions of emotion.
\newblock {\em PNAS}, 2014.

\bibitem{esser2018variational}
P.~Esser, E.~Sutter, and B.~Ommer.
\newblock A variational u-net for conditional appearance and shape generation.
\newblock In {\em CVPR}, 2018.

\bibitem{Goodfellow2014Generative}
I.~J. Goodfellow, J.~Pougetabadie, M.~Mirza, B.~Xu, D.~Wardefarley, S.~Ozair,
  A.~Courville, and Y.~Bengio.
\newblock Generative adversarial networks.
\newblock In {\em NIPS}, 2014.

\bibitem{gulrajani2017improved}
I.~Gulrajani, F.~Ahmed, M.~Arjovsky, V.~Dumoulin, and A.~C. Courville.
\newblock Improved training of wasserstein gans.
\newblock In {\em NIPS}, 2017.

\bibitem{he2016deep}
K.~He, X.~Zhang, S.~Ren, and J.~Sun.
\newblock Deep residual learning for image recognition.
\newblock In {\em CVPR}, 2016.

\bibitem{he2019attgan}
Z.~He, W.~Zuo, M.~Kan, S.~Shan, and X.~Chen.
\newblock Attgan: Facial attribute editing by only changing what you want.
\newblock {\em IEEE Transactions on Image Processing}, 2019.

\bibitem{heusel2017gans}
M.~Heusel, H.~Ramsauer, T.~Unterthiner, B.~Nessler, and S.~Hochreiter.
\newblock Gans trained by a two time-scale update rule converge to a local nash
  equilibrium.
\newblock In {\em NIPS}, 2017.

\bibitem{isola2017image}
P.~Isola, J.-Y. Zhu, T.~Zhou, and A.~A. Efros.
\newblock Image-to-image translation with conditional adversarial networks.
\newblock In {\em CVPR}. IEEE, 2017.

\bibitem{Kemelmacher2014Illumination}
I.~Kemelmacher-Shlizerman, S.~Suwajanakorn, and S.~M. Seitz.
\newblock Illumination-aware age progression.
\newblock In {\em CVPR}, 2014.

\bibitem{lample2017unsupervised}
G.~Lample, A.~Conneau, L.~Denoyer, and M.~Ranzato.
\newblock Unsupervised machine translation using monolingual corpora only.
\newblock {\em ICLR}, 2018.

\bibitem{ledig2017photo}
C.~Ledig, L.~Theis, F.~Husz{\'a}r, J.~Caballero, A.~Cunningham, A.~Acosta,
  A.~P. Aitken, A.~Tejani, J.~Totz, Z.~Wang, et~al.
\newblock Photo-realistic single image super-resolution using a generative
  adversarial network.
\newblock In {\em CVPR}, 2017.

\bibitem{li2016deep}
M.~Li, W.~Zuo, and D.~Zhang.
\newblock Deep identity-aware transfer of facial attributes.
\newblock {\em arXiv:1610.05586}, 2016.

\bibitem{li2018beautygan}
T.~Li, R.~Qian, C.~Dong, S.~Liu, Q.~Yan, W.~Zhu, and L.~Lin.
\newblock Beautygan: Instance-level facial makeup transfer with deep generative
  adversarial network.
\newblock In {\em MM}, 2018.

\bibitem{liu2017face}
S.~Liu, Y.~Sun, D.~Zhu, R.~Bao, W.~Wang, X.~Shu, and S.~Yan.
\newblock Face aging with contextual generative adversarial nets.
\newblock In {\em MM}, 2017.

\bibitem{liu2018cross}
S.~Liu, Y.~Sun, D.~Zhu, G.~Ren, Y.~Chen, J.~Feng, and J.~Han.
\newblock Cross-domain human parsing via adversarial feature and label
  adaptation.
\newblock {\em arXiv:1801.01260}, 2018.

\bibitem{mirza2014conditional}
M.~Mirza and S.~Osindero.
\newblock Conditional generative adversarial nets.
\newblock {\em arXiv:1411.1784}, 2014.

\bibitem{miyato2018spectral}
T.~Miyato, T.~Kataoka, M.~Koyama, and Y.~Yoshida.
\newblock Spectral normalization for generative adversarial networks.
\newblock {\em arXiv:1802.05957}, 2018.

\bibitem{miyato2018cgans}
T.~Miyato and M.~Koyama.
\newblock cgans with projection discriminator.
\newblock {\em arXiv:1802.05637}, 2018.

\bibitem{perarnau2016invertible}
G.~Perarnau, J.~van~de Weijer, B.~Raducanu, and J.~M. {\'A}lvarez.
\newblock Invertible conditional gans for image editing.
\newblock {\em arXiv:1611.06355}, 2016.

\bibitem{pumarola2018ganimation}
A.~Pumarola, A.~Agudo, A.~M. Martinez, A.~Sanfeliu, and F.~Moreno-Noguer.
\newblock Ganimation: Anatomically-aware facial animation from a single image.
\newblock In {\em ECCV}, 2018.

\bibitem{shrivastava2017learning}
A.~Shrivastava, T.~Pfister, O.~Tuzel, J.~Susskind, W.~Wang, and R.~Webb.
\newblock Learning from simulated and unsupervised images through adversarial
  training.
\newblock In {\em CVPR}, 2017.

\bibitem{vaswani2017attention}
A.~Vaswani, N.~Shazeer, N.~Parmar, J.~Uszkoreit, L.~Jones, A.~N. Gomez,
  {\L}.~Kaiser, and I.~Polosukhin.
\newblock Attention is all you need.
\newblock In {\em NIPS}, 2017.

\bibitem{wang2017high}
T.-C. Wang, M.-Y. Liu, J.-Y. Zhu, A.~Tao, J.~Kautz, and B.~Catanzaro.
\newblock High-resolution image synthesis and semantic manipulation with
  conditional gans.
\newblock {\em arXiv:1711.11585}, 2017.

\bibitem{wang2018non}
X.~Wang, R.~Girshick, A.~Gupta, and K.~He.
\newblock Non-local neural networks.
\newblock In {\em CVPR}, 2018.

\bibitem{yao20183d}
S.~Yao, T.~M.~H. Hsu, J.-Y. Zhu, J.~Wu, A.~Torralba, B.~Freeman, and
  J.~Tenenbaum.
\newblock 3d-aware scene manipulation via inverse graphics.
\newblock {\em arXiv:1808.09351}, 2018.

\bibitem{zhang2018generative}
G.~Zhang, M.~Kan, S.~Shan, and X.~Chen.
\newblock Generative adversarial network with spatial attention for face
  attribute editing.
\newblock In {\em ECCV}, 2018.

\bibitem{zhang2017stackgan}
H.~Zhang, T.~Xu, H.~Li, S.~Zhang, X.~Huang, X.~Wang, and D.~Metaxas.
\newblock Stackgan: Text to photo-realistic image synthesis with stacked
  generative adversarial networks.
\newblock {\em arXiv preprint}, 2017.

\bibitem{zhang2017age}
Z.~Zhang, Y.~Song, and H.~Qi.
\newblock Age progression/regression by conditional adversarial autoencoder.
\newblock In {\em CVPR}, 2017.

\bibitem{zhao2018modular}
B.~Zhao, B.~Chang, Z.~Jie, and L.~Sigal.
\newblock Modular generative adversarial networks.
\newblock In {\em ECCV}, 2018.

\bibitem{Zhu2017Unpaired}
J.-Y. Zhu, T.~Park, P.~Isola, and A.~A. Efros.
\newblock Unpaired image-to-image translation using cycle-consistent
  adversarial networks.
\newblock {\em arXiv:1703.10593}, 2017.

\bibitem{zhu2017toward}
J.-Y. Zhu, R.~Zhang, D.~Pathak, T.~Darrell, A.~A. Efros, O.~Wang, and
  E.~Shechtman.
\newblock Toward multimodal image-to-image translation.
\newblock In {\em NIPS}, 2017.

\end{thebibliography}
}   

\newpage

 \onecolumn
\begin{appendices}
\section{Network Architecture}
\begin{figure*}[h]
	\begin{center}
		\includegraphics[width=0.5\textwidth]{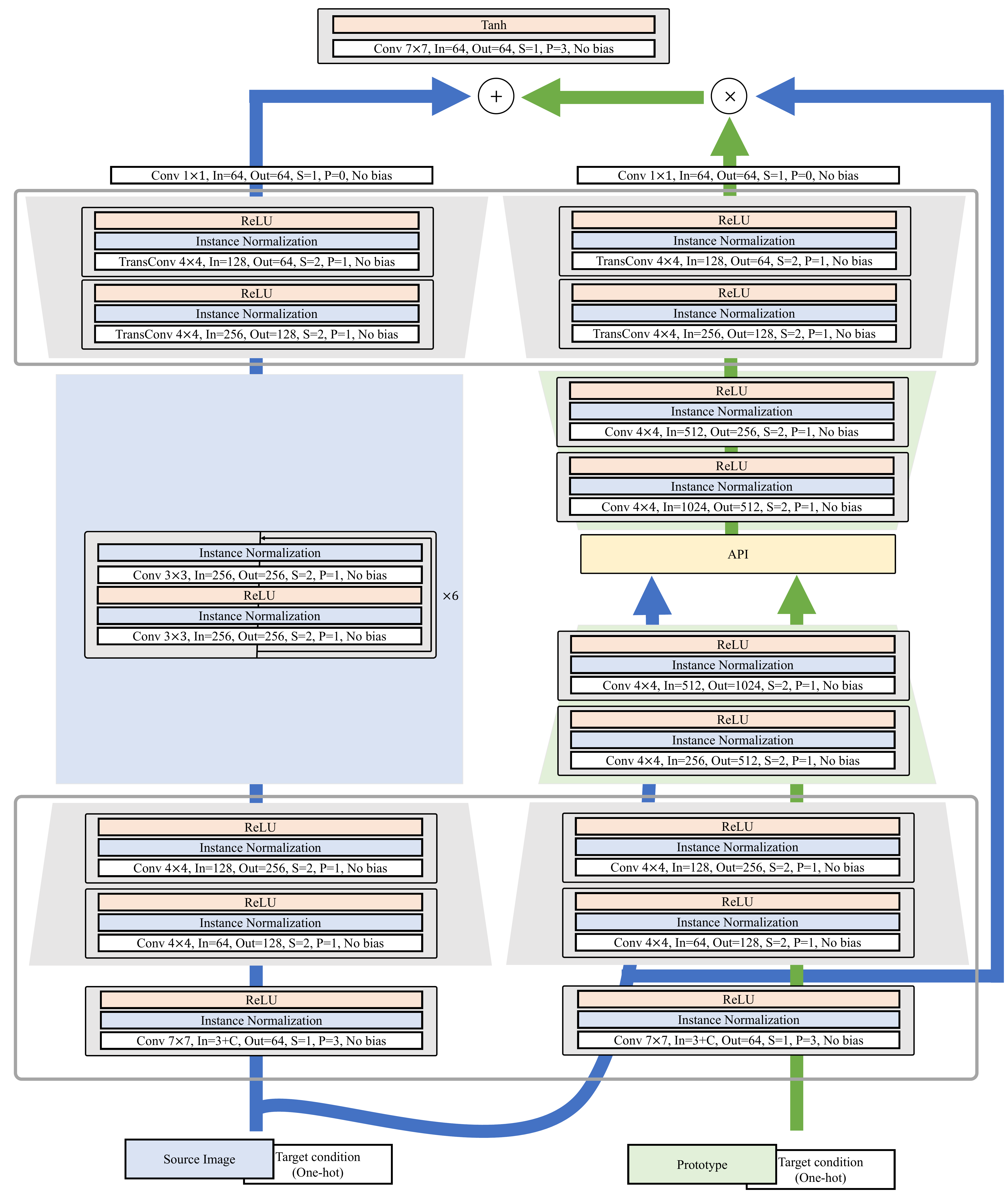}
	\end{center}
	\caption{The architecture of the generator.}
	\label{fig:G_architecture}
\end{figure*}

\begin{figure*}[h]
	\begin{center}
		\includegraphics[width=0.5\textwidth]{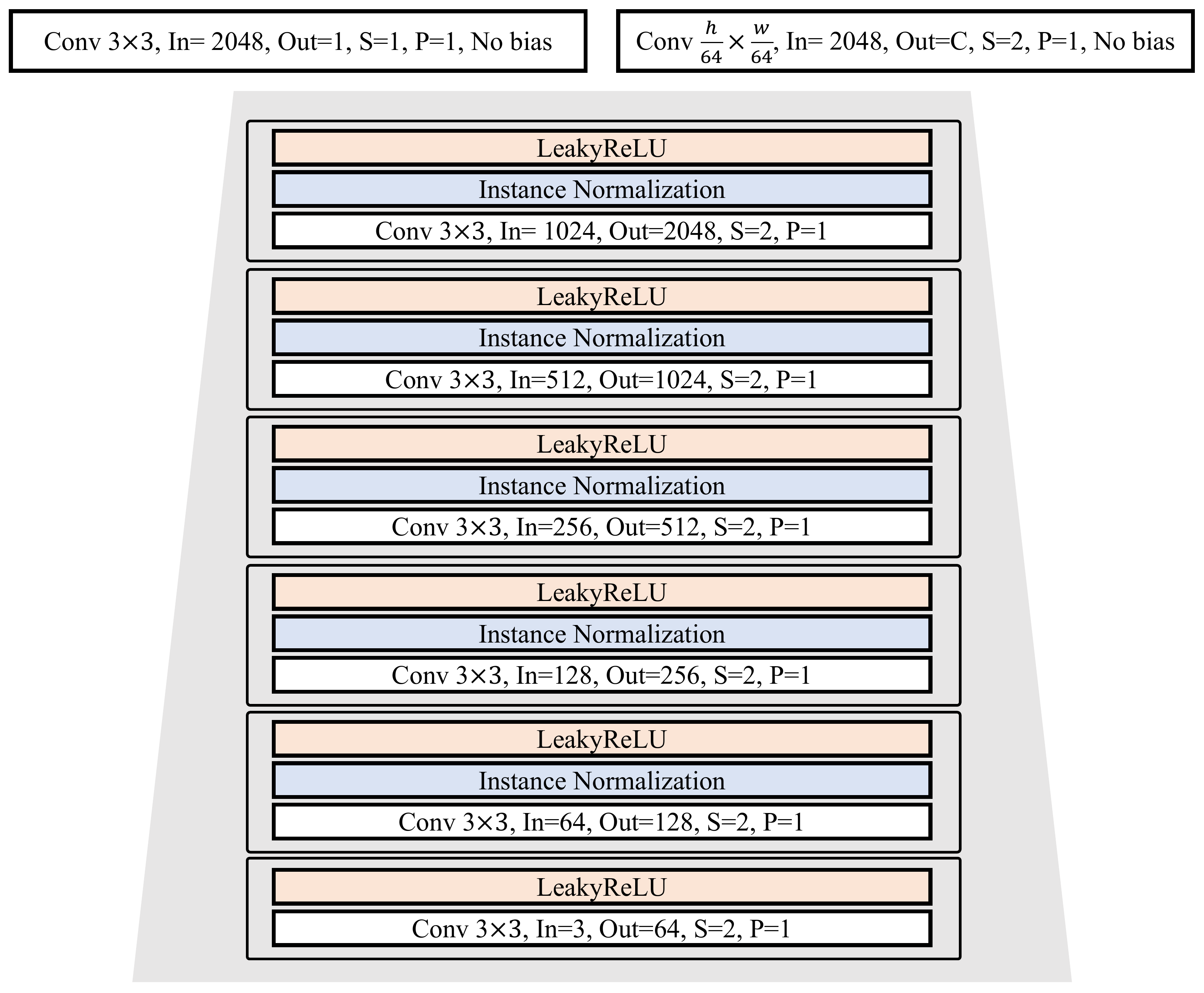}
	\end{center}
	\caption{The architecture of the discriminator.}
	\label{fig:D_architecture}
\end{figure*}

The architectures of the generator and discriminator are shown in Figure. \ref{fig:G_architecture} and \ref{fig:D_architecture}. 

\section{Prototype}
\begin{figure*}[h]
	\begin{center}
		\includegraphics[width=1\textwidth]{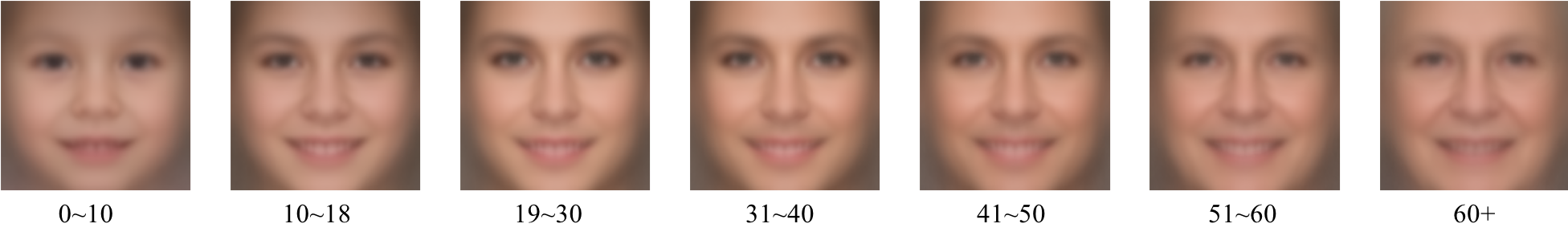}
	\end{center}

	\caption{Average images of face aging dataset.}

	\label{fig:ave_age}
\end{figure*}

\begin{figure*}[h]
	\begin{center}
		\includegraphics[width=1\textwidth]{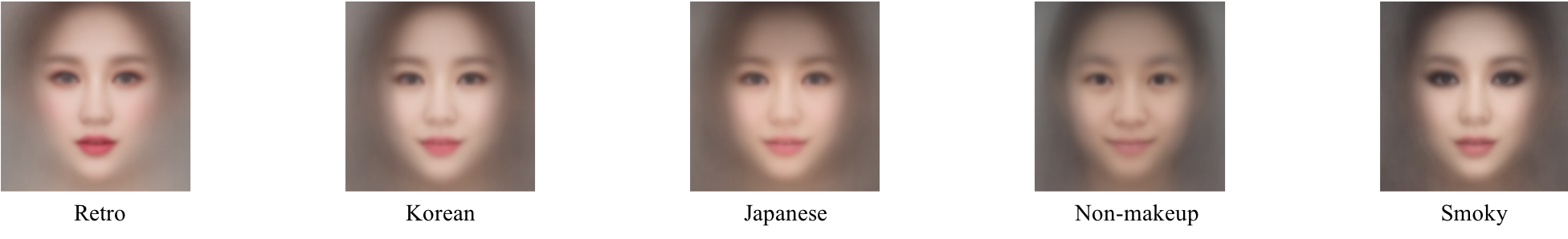}
	\end{center}

	\caption{Average images of MAKEUP-A5 dataset.}

	\label{fig:ave_makeup}
\end{figure*}

\begin{figure*}[h]
	\begin{center}
		\includegraphics[width=1\textwidth]{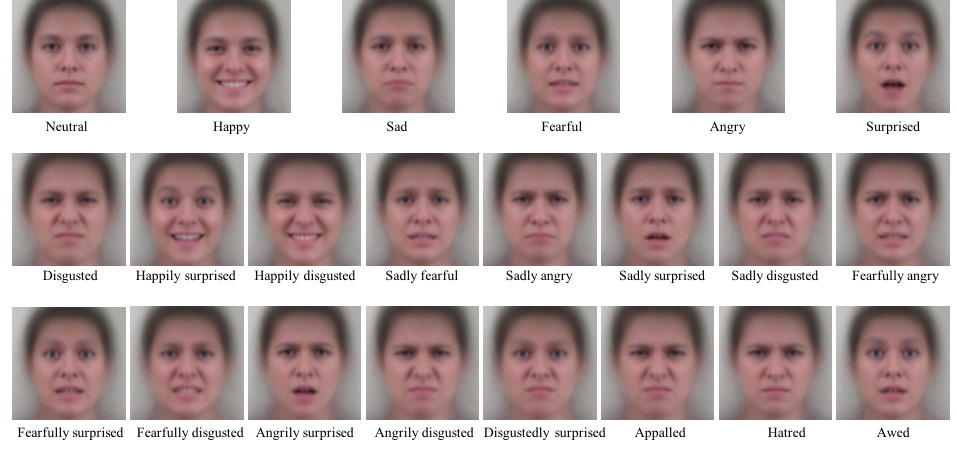}
	\end{center}

	\caption{Average images of CFEE dataset.}

	\label{fig:ave_cfee}
\end{figure*}

We take the average image of as the prototype of each domain. The average images of the datasets are shown in Figure \ref{fig:ave_age}, \ref{fig:ave_makeup}, \ref{fig:ave_cfee}.  

\section{Qualitative Results }
\begin{figure*}[t]
	\begin{center}
		\includegraphics[width=1\textwidth]{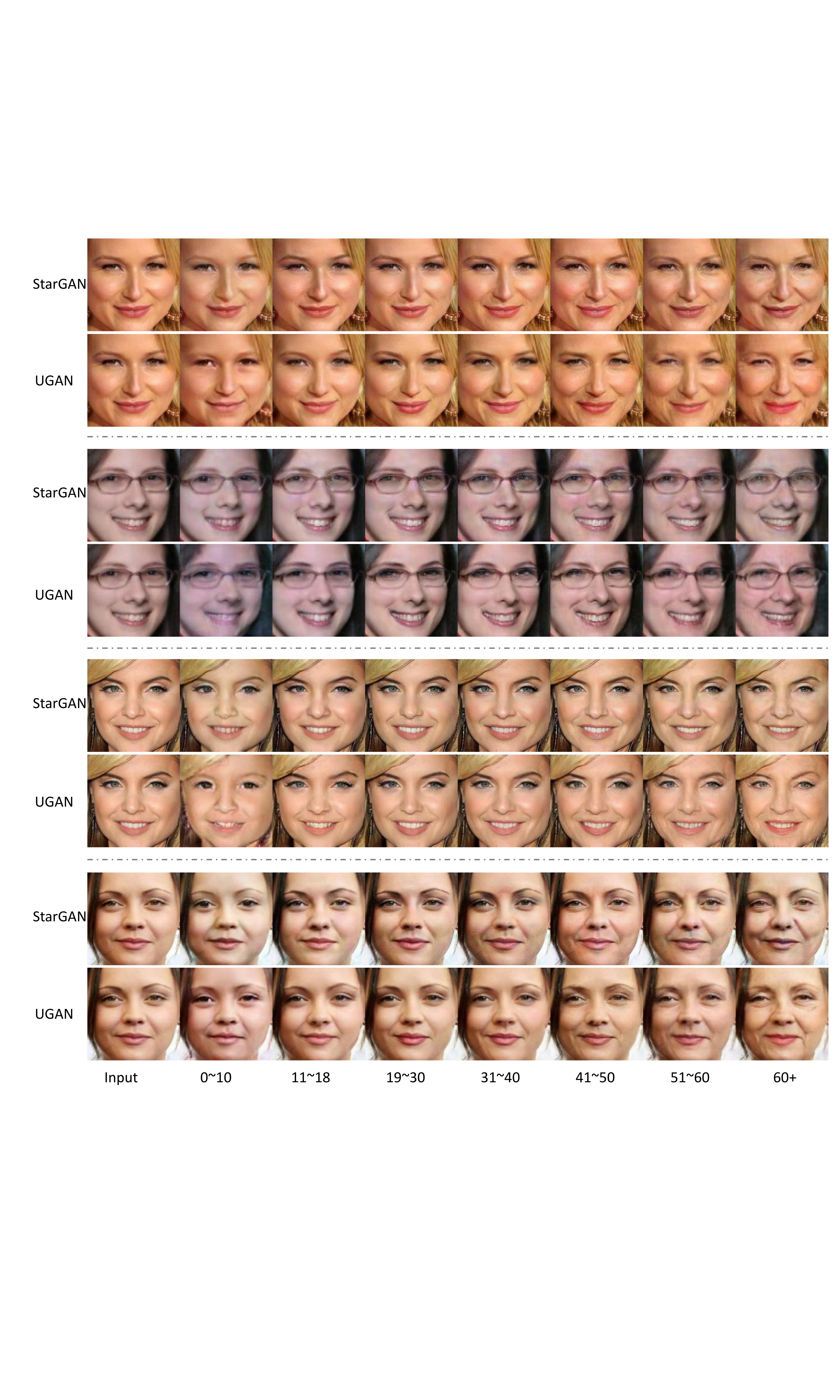}
	\end{center}

	\caption{Face aging results on the face aging dataset.}

	\label{fig:aging_1}
\end{figure*}

\begin{figure*}[t]
	\begin{center}
		\includegraphics[width=1\textwidth]{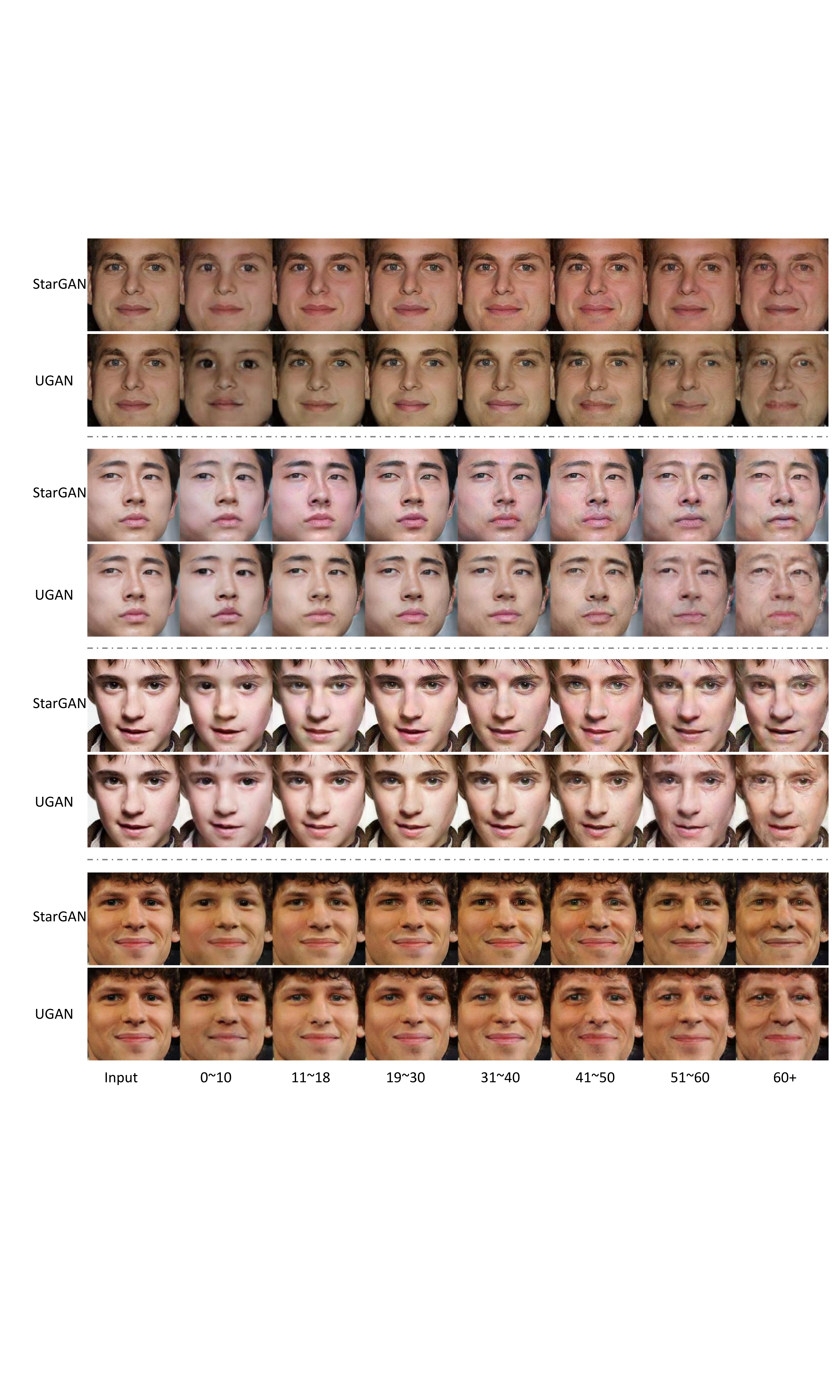}
	\end{center}

	\caption{Face aging results on the face aging dataset.}

	\label{fig:aging_2}
\end{figure*}

\begin{figure*}[t]
	\begin{center}
		\includegraphics[width=0.9\textwidth]{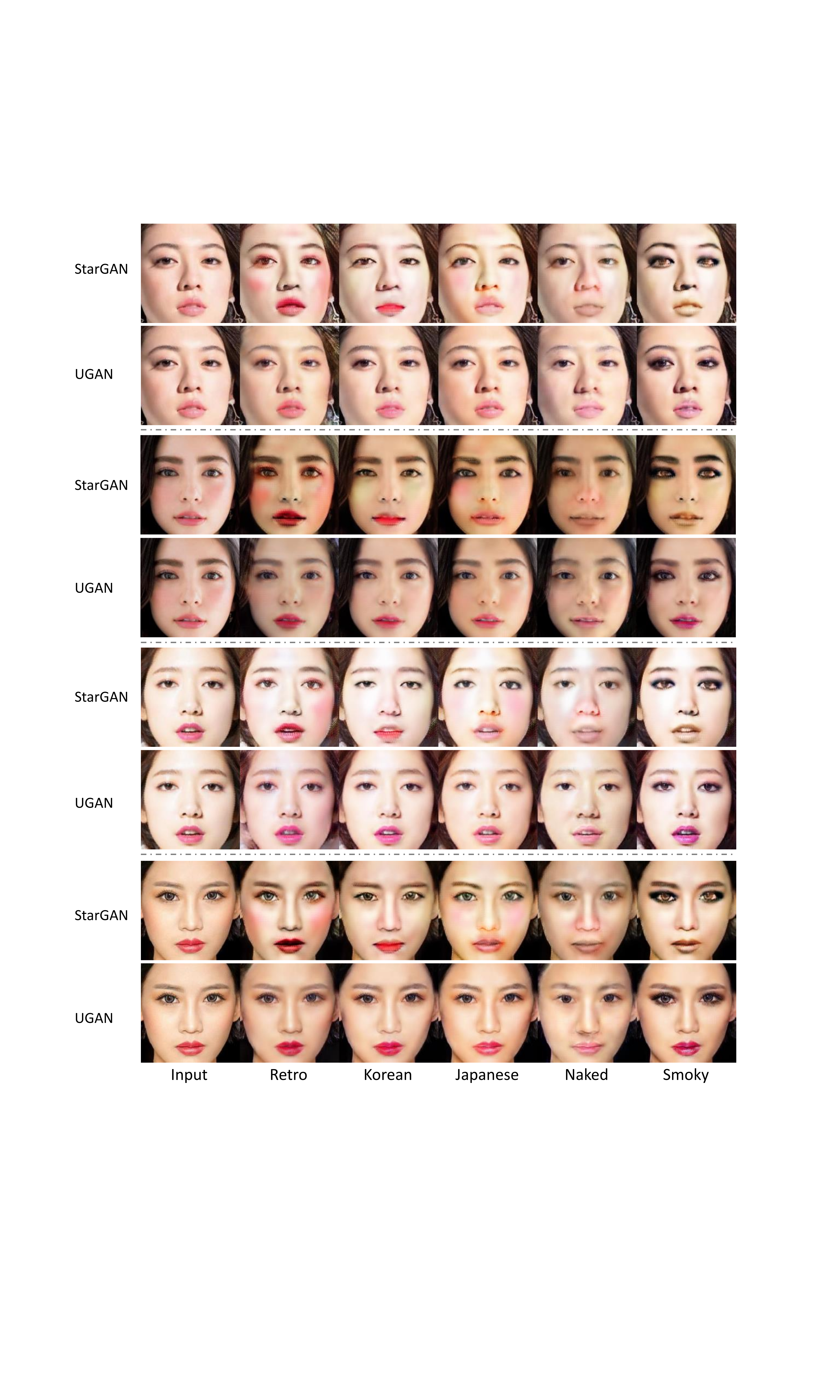}
	\end{center}

	\caption{Makeup editing results on the MAKEUP-A5 dataset.}

	\label{fig:makeup_1}
\end{figure*}

\begin{figure*}[t]
	\begin{center}
		\includegraphics[width=0.9\textwidth]{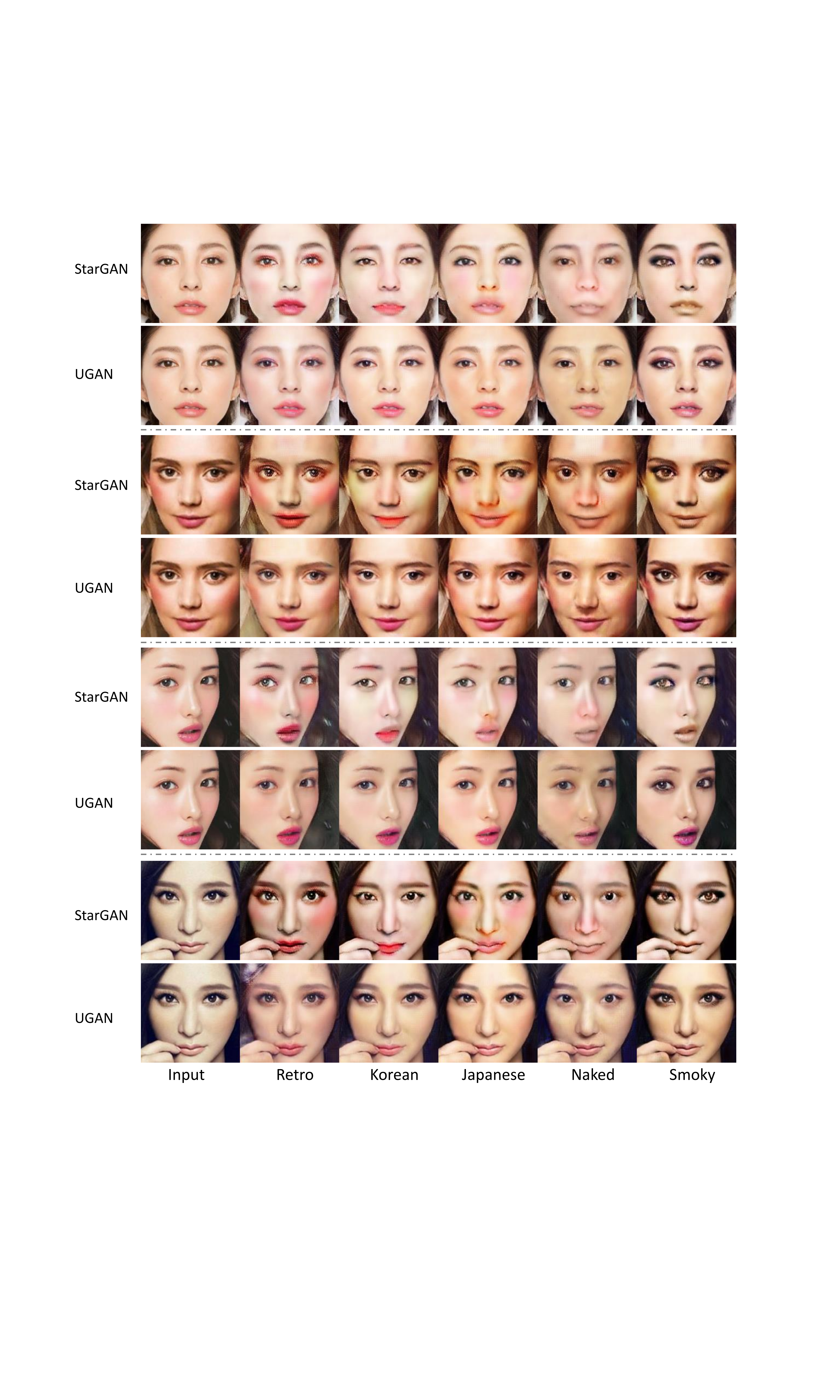}
	\end{center}
	\vspace{-4mm}
	\caption{Makeup editing results on the MAKEUP-A5 dataset.}
	\vspace{-4mm}
	\label{fig:makeup_2}
\end{figure*}

\begin{figure*}[t]
	\begin{center}
		\includegraphics[width=0.75\textwidth]{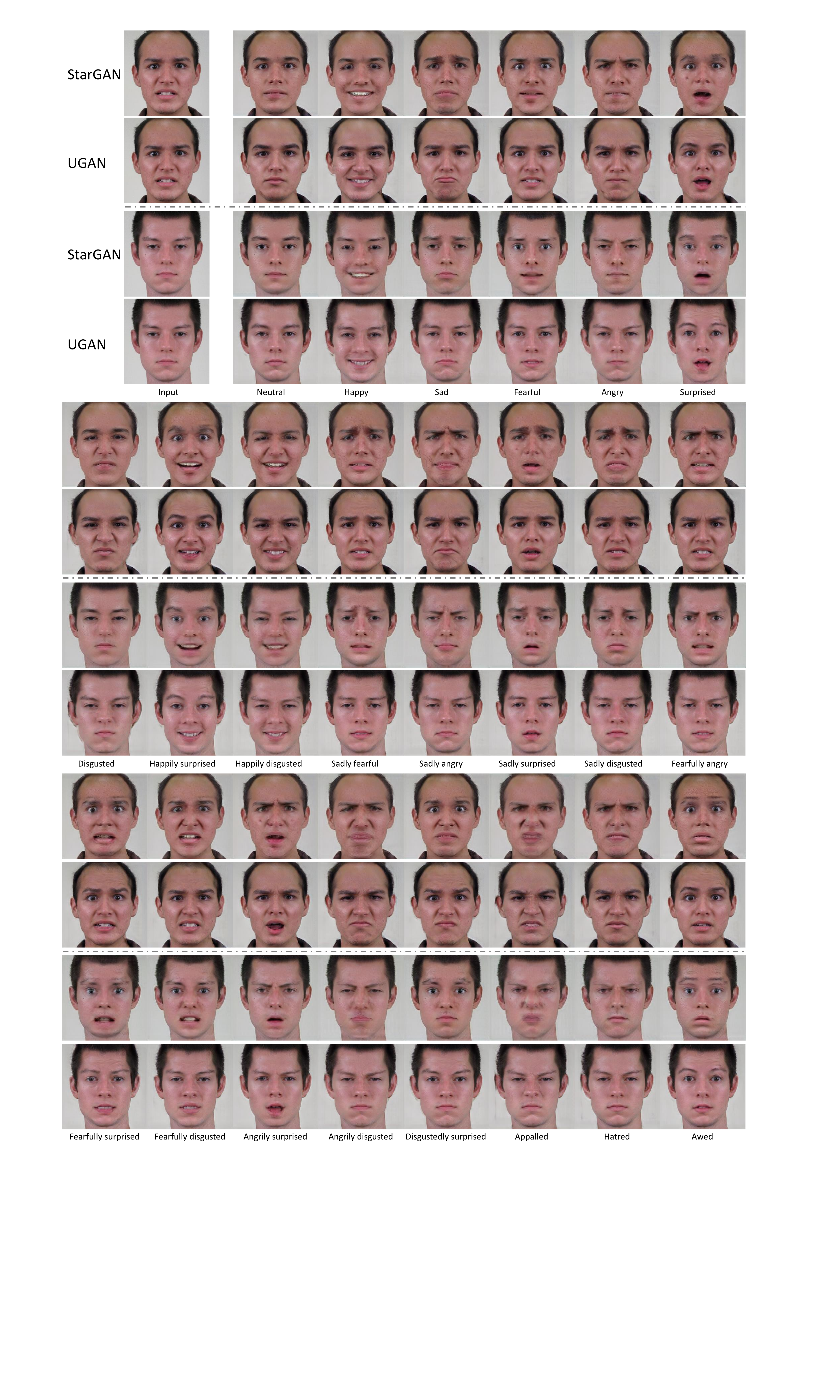}
	\end{center}

	\caption{Expression editing results on the CFEE dataset.}

	\label{fig:expression_1}
\end{figure*}

\begin{figure*}[h]
	\begin{center}
		\includegraphics[width=0.75\textwidth]{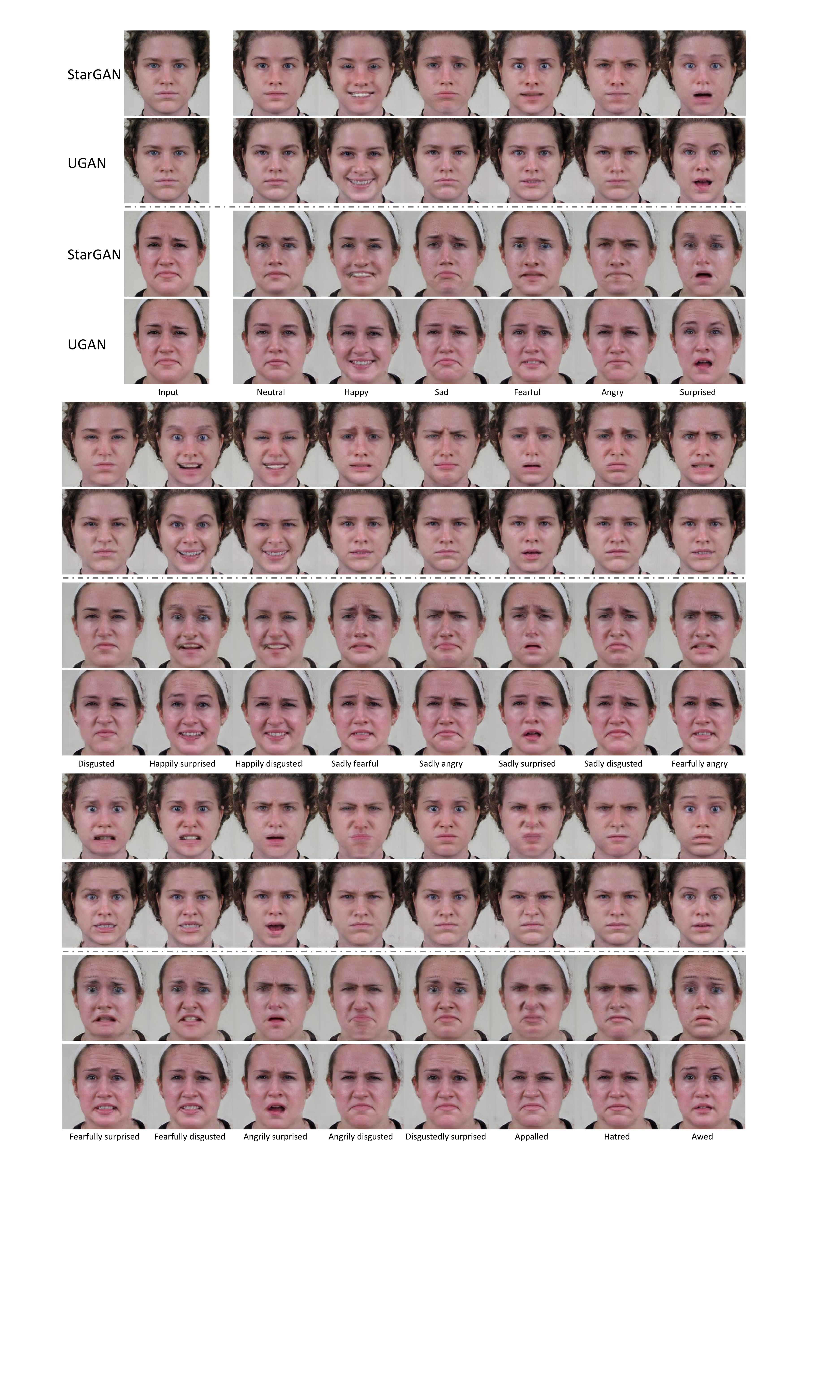}
	\end{center}
	\caption{Expression editing results on the CFEE dataset.}
	\label{fig:expression_2}
\end{figure*}

The face editing results on face aging, MAKEUP-A5 (makeup editing) and CFEE (expression editing) datasets are shown in Figure \ref{fig:aging_1}, \ref{fig:aging_2}, \ref{fig:makeup_1}, \ref{fig:makeup_2}, \ref{fig:expression_1} and \ref{fig:expression_2}.

{\bf Face aging}: the results on face aging dataset are shown in Figure \ref{fig:aging_1} and \ref{fig:aging_2}. In the Figure \ref{fig:aging_1}, the women images are used as input and synthesized faces of seven age groups are shown in second to seventh columns. 
Observing the second and last columns, our method generates very realistic results. 
For example, in the sixth row and fourth column $\langle row \; 6, col\; 4 \rangle$ of Figure \ref{fig:aging_1}, the woman is successfully transformed into a child with baby teeth, big eyes, etc. 
For another example, in $\langle row\; 4, col \;8 \rangle$ the woman is aged to a senior woman with white hair and wrinkles.
Similar conclusions can be drawn by taking men as input as shown in \ref{fig:aging_2}. 
For example, in $\langle row \;2, col \;4 \sim 6 \rangle$, the beard of the translated images become increasingly thicker.

{\bf Makeup editing}: $4$ exemplar results of StarGAN and UGAN on MAKEUP-A5 are displayed in Figure \ref{fig:makeup_1} and \ref{fig:makeup_2} respectively. Observing the images of the fifth column, all makeup can be removed to be a naked face. 
By observing the others columns, the makeup results of our method correspond to the specified categories.
For example, in $\langle row \;8, col \; 2 \rangle$ of Figure \ref{fig:makeup_1}, the translated face belongs to ``Retro'' with pink blush, lipstick, eye shadow. 
For another example, in $\langle row \;8, col \; 6 \rangle$ of Figure \ref{fig:makeup_1}, the translated face belongs to ``Smoky'' with black eyeliner and eye shadow.

{\bf Expression editing}: $2$ exemplar results of expression editing on CFEE are demonstrated in Figure \ref{fig:expression_1} and \ref{fig:expression_2} respectively. Our method is able to edit $22$ kinds of fine-grained facial expression well.
For example, for the image in the second row of Figure \ref{fig:expression_2}, when translating it to ``happy'', our method successfully synthesizes the real teeth and accurately expresses the happy expression.
Our method also can vividly synthesize other expressions. 

\end{appendices}

\end{document}